\documentclass{article} 
\usepackage[accepted]{icml2019_arxiv}
\usepackage{hyperref}
\usepackage{url}
\usepackage{amsthm}
\usepackage{amssymb,dsfont,amsmath,amsthm,mathtools,natbib}
\usepackage{enumitem}

\newtheorem{property}{Property}
\newtheorem{example}{Example}
\newtheorem{remark}{Remark}

\usepackage{xspace}
\newcommand{\pwl}{Proximal Weighted \emph{Lasso}\xspace}

\def\R{\mathbb{R}}

\def\x{\mathbf{x}}
\def\X{\mathbf{X}}

\def\v{\mathbf{v}}
\def\y{\mathbf{y}}
\def\r{\mathbf{r}}

\def\a{\mathbf{a}}
\def\e{\mathbf{e}}

\def\s{\mathbf{s}}

\def\I{\mathbf{I}}

\def\w{\mathbf{w}}

\def\cE{\mathcal{E}}

\def\bfeta{\boldsymbol{\eta}}

\DeclareMathOperator{\sign}{sign}

\newcommand{\argmin}{\mathop{\mathrm{arg\,min}}}
\newcommand{\argmax}{\mathop{\mathrm{arg\,max}}}

\DeclareMathOperator{\newexp}{\nu}

\newcommand{\eg}{{\em e.g.,~}}

\newcommand{\wrt}{{\em w.r.t.~}}

\usepackage[colorinlistoftodos,bordercolor=orange,backgroundcolor=orange!20,linecolor=orange,textsize=scriptsize]{todonotes}

\usepackage{cleveref}
\begin{document}

\twocolumn[

\icmltitle{Screening Rules for Lasso with  Non-Convex Sparse Regularizers}

\icmlsetsymbol{equal}{*}

\begin{icmlauthorlist}
	\icmlauthor{Alain Rakotomamonjy}{to}
	\icmlauthor{Gilles Gasso}{goo}
	\icmlauthor{Joseph Salmon}{ed}
\end{icmlauthorlist}

\icmlaffiliation{to}{Universit\'e Rouen Normandie, Criteo AI Lab}
\icmlaffiliation{goo}{INSA Rouen Normandie}
\icmlaffiliation{ed}{Université\'e de Montpellier}
\icmlcorrespondingauthor{alain.rakotomamonjy}{first.last@insa-rouen.fr}

\icmlkeywords{dsd}

\vskip 0.3in
]
\printAffiliationsAndNotice{} 
\begin{abstract}
	Leveraging on the convexity of the Lasso problem, screening rules
	help in accelerating solvers by discarding irrelevant variables, during the optimization process. However, because
	they provide better theoretical guarantees in identifying relevant
	variables, several non-convex regularizers for the Lasso have been proposed in the literature.  This work is the first that introduces
	a screening rule strategy into a non-convex Lasso solver. The approach we propose is based on a iterative majorization-minimization (MM) strategy that includes a screening rule in the  inner solver and a condition for propagating screened variables between iterations of MM. In addition to improve efficiency of solvers, we also provide guarantees that the inner solver is able to identify the zeros components of its critical point in finite time. 
	Our experimental analysis illustrates
	the significant computational gain  brought by the new screening rule compared to classical coordinate-descent  or proximal gradient descent methods.
\end{abstract}

\section{Introduction}
\label{sec:introduction}

Sparsity-inducing penalties are classical tools in statistical machine learning, especially in settings where the data available for learning is
scarce and of high-dimension.
In addition, when the solution of the learning problem is known to be sparse, using those penalties yield to models that can leverage this prior knowledge. The \emph{Lasso} \citep{tibshirani1996regression} and the \emph{Basis pursuit} \citep{chen2001atomic,chenbasispursuit} where
the first approaches that have employed $\ell_1$-norm penalty for inducing
sparsity.

While the \emph{Lasso} has had a great impact on machine learning and signal processing communities given some success stories \citep{shevade2003simple,donoho2006compressed,lustig2008compressed,ye2012sparse}
it also comes with some theoretical drawbacks (\eg biased estimates of large
coefficient of the model).
Hence, several authors have proposed non-convex penalties that approximate better the $\ell_0$-(pseudo)norm, the later being the original measure of sparsity though it leads to NP-hard learning problem.
The most commonly used non-convex penalties are the \emph{Smoothly Clipped
Absolute Deviation} (SCAD) \cite{fan2001variable}, the \emph{Log Sum penalty} (LSP) \cite{candes2008enhancing}, the \emph{capped-$\ell_1$ penalty} \cite{zhang2010analysis}, the \emph{Minimax Concave Penalty} (MCP) \cite{zhang2010nearly}.
We refer the interested reader to \citep{Soubies_Blanc-FeraudAubert16} for a discussion on the pros and cons of such non-convex formulations.

From an optimization point of view, the \emph{Lasso} can benefit from
a large palette of algorithms ranging from block-coordinate descent
\cite{friedman2007pathwise,fu1998penalized}
to (accelerated) proximal gradient approaches \cite{beck2009fast}.
In addition, by leveraging convex optimization theory, some of these algorithms can be further accelerated by combining them with sequential or dynamic safe screening rules \cite{ghaouisafe2012,bonnefoy2015dynamic,fercoq2015gap}, which allow to safely set some useless variables to $0$ before terminating (or even sometimes before starting) the algorithm.

While learning sparse models with non-convex penalties are seemingly more
challenging to address, block-coordinate descent algorithms \citep{breheny2011coordinate,mazumder2011sparsenet} or iterative shrinkage thresholding \citep{gong2013general} algorithms can be extended to those learning problems.
Another popular way of handling non-convexity is to consider the majorization-minimization (MM) \cite{hunter2004tutorial} principle which consists in iteratively minimizing a majorization of a (non-convex) objective function.
When applied to non-convex sparsity enforcing penalty, the MM scheme leads to solving a sequence of weighted \emph{Lasso} problems \cite{candes2008enhancing, gasso2009recovering,mairal2013stochastic}.

In this paper, we propose a screening strategy that can be applied when dealing with non-convex penalties.
As far as we know, this work is the first attempt in that direction.
For that, we consider a MM framework and in this context, our contributions are
\begin{itemize}
\item the definition of a MM algorithm that produces a sequence of iterates known to converge towards a critical point of our non-convex learning problem,
\item the proposition of a duality gap based screening rule for weighted \emph{Lasso}, which is the core algorithm of our MM framework,
\item  the introduction of conditions allowing to propagate screened variables from one MM iteration to the next,
\item we also empirically show that our screening strategy indeed improves the running time of MM algorithms with respect to block-coordinate descent or proximal gradient descent methods.
\end{itemize}

\section{Global non-convex framework}
\label{sec:framework_an_mm_algorithm}
We introduce the problem we are interesting in, its first order optimality conditions and propose an MM approach for its resolution.

\subsection{The optimization problem}
\label{sub:the_optimization_problem}

We consider solving the problem of least-squares regression with a generic penalty of the form
\begin{equation}\label{eq:generalprob}
	\min_{\w\in \R^d} \frac{1}{2} \| \y - \X\w\|_2^2 + \sum_{j=1}^d r_\lambda(|w_j|) \enspace,
\end{equation}
where $\y \in \R^n$ is a target vector, $\X=[\x_1,\dots,\x_d] \in \R^{n \times d}$ is the design matrix with column-wise features $\x_j$, $\w$ is the coefficient vector of the model and the map $r_\lambda: \R_+ \mapsto \R_+$ is concave and differentiable on $[0,+\infty)$ with a regularization parameter $\lambda > 0$.
In addition, we assume that $r_{\lambda}(|w|)$ is lower semi-continuous function.
Note that most penalty functions such as SCAD, MCP or log sum (see their definitions in Table \ref{tab:ncvx_pen}) admit such a property.

\begin{table*}[t]
\caption{Common non-convex penalties  with their sub-differentials. Here $\lambda > 0$, $\theta > 0$ ($\theta > 1$ for MCP, $\theta > 2$ for SCAD).
}
\label{tab:ncvx_pen}
\vskip 0.15in
\begin{center}
	\begin{small}
				\begin{tabular}{l | l | l}
			\hline
			Penalty & $r_\lambda(|w|)$ &  $\partial r_\lambda(|w|)$ \\
			\hline
			Log sum    & $\lambda \log(1 + |w|/\theta)$ & $ \displaystyle \left\lbrace
			\begin{array}{lll}
			\left[\frac{-\lambda}{\theta}, \frac{\lambda}{\theta} \right] & \text{if} & w = 0  \\
			\left \{\lambda \frac{\sign(w)}{\theta + |w|} \right \} & \text{if} & w > 0
			\end{array} \right.$
			\\
						\hline
			MCP & $\left\lbrace
			\begin{array}{lll}
			\lambda |w| - \frac{w^2}{2 \theta} & \text{if} & |w| \leq \lambda \theta \\
			\theta \lambda^2/2 & \text{if} & |w| > \theta \lambda
			\end{array} \right.$
			& $\left\lbrace \begin{array}{lll}
			\left[-\lambda, \lambda \right] & \text{if} & w = 0 \\
			\{\lambda \sign(w) - \frac{w}{\theta}\} & \text{if} & 0 < |w| \leq \lambda \theta \\
			\{0 \} & \text{if} & |w| > \theta \lambda
			\end{array} \right.$
			\\
			\hline

			SCAD & $\left\lbrace
			\begin{array}{lll}
			\lambda |w|  & \text{if} & |w| \leq \lambda  \\
			\frac{1}{2 (\theta - 1)} (- w^2 + 2 \theta \lambda |w| - \lambda^2 ) & \text{if} & \lambda < |w| \leq \lambda \theta \\
			\frac{\lambda^2 (1+\theta)}{2} & \text{if} & |w| > \theta \lambda
			\end{array} \right.$
			&  $\left\lbrace \begin{array}{lll}
			\left[-\lambda, \lambda \right] & \text{if} & w = 0 \\
			\{\lambda \sign(w)\} & \text{if} & 0 < |w| \leq \lambda \\
			\left\{\frac{1}{\theta - 1}(- w +  \theta \lambda \sign(w)  )\right\} & \text{if} & 0 < |w| \leq \lambda \theta \\
			\{0 \} & \text{if} & |w| > \theta \lambda
			\end{array} \right.$  \\
			\hline
		\end{tabular}
			\end{small}
\end{center}
\vskip -0.1in
\end{table*}

We consider tools such as Fr\'echet subdifferentials and limiting-subdifferentials \citep{kruger2003frechet,rockafellar2009variational,mordukhovich2006frechet} well suited for non-smooth and non-convex optimization, so that a vector $\w^\star$ belongs to the set of minimizers (not necessarily global) of Problem \eqref{eq:generalprob} if the following Fermat condition holds \citep{Clarke89,kruger2003frechet}:
\begin{equation}\label{eq:fermat}
   \X^\top (\y - \X \w^\star) \in \sum_{j=1}^d\partial r_{\lambda}(|w_j^\star|) \enspace,
\end{equation}
with $\partial r_\lambda(\cdot)$ being the Fr\'echet subdifferential of $r_{\lambda}$, assuming it exists at $\w^\star$. In particular this is the case for the MCP, log sum and SCAD penalties presented in \Cref{tab:ncvx_pen}.
For an illustration, we present the optimality conditions for MCP and log sum.

\begin{example} For the MCP penalty (see Table \ref{tab:ncvx_pen} for the definition and subdifferential), it is easy to show that the $\partial r_{\lambda}(0)= [-\lambda, \lambda]$. Hence,
	 the Fermat condition becomes
	\begin{align}\label{eq:optcondmcp}
		- \x_j^\top (\y - \X \w^\star)  = 0, \quad & \text{~if~} |w_j^\star| > \lambda \theta \nonumber\\
		|\x_j^\top (\y - \X \w^\star)| \leq {\lambda},  \quad & \text{~if~} w_j^\star = 0 \nonumber	\\
			- \x_j^\top (\y - \X \w^\star)  + \lambda \sign(w_j^\star) - \tfrac{w_j^\star}{\theta}
					= 0, \,\, & \text{~otherwise~} \enspace.
\end{align}
\end{example}
\begin{example} For the log sum penalty
one can explicitly compute
$\partial r_{\lambda}(0) = [-\frac{\lambda}{\theta},\frac{\lambda}{\theta}]$ and leverage on smoothness of $r_{\lambda}(|w|)$
when $|w|>0$ for computing $\partial r_{\lambda}(|w|)$. Then the above necessary condition can be translated as
\begin{align}\label{eq:optcond}
 - \x_j^\top (\y - \X \w^\star)  + \lambda \frac{\sign(w_j^\star)}{\theta + |w_j^\star|} = 0, \quad & \text{if~} w_j^\star \neq 0 \enspace,\nonumber\\
|\x_j^\top (\y - \X \w^\star)| \leq \frac{\lambda}{\theta},  \quad & \text{if~} w_j^\star = 0 \enspace.
\end{align}
\end{example}
As we can see, first order optimality conditions lead to simple equations and inclusion.
\begin{remark}
There exists a critical parameter $\lambda_{\max}$ such that $\mathbf{0}$ is a critical point for the primal problem for all $\lambda \geq
\lambda_{\max}$.
This parameter depends on the subdifferential of $r_{\lambda}$ at $0$.
For the MCP penalty we have $\lambda_{\max} \triangleq  \max_j{|\x_j^\top \y|}$, and for the log sum penalty $\lambda_{\max} \triangleq \theta \max_j{|\x_j^\top \y|}$.
From now on, we assume that $\lambda \leq \lambda_{\max}$ to avoid such irrelevant local solutions.
\end{remark}

\subsection{Majorization-Minimization approach}
\label{sub:majorization_minimization_approach}

There exists several majorization-minimization algorithms for solving non-smooth and non-convex problems involving sparsity-inducing penalties \cite{gasso2009recovering,gong2013general,mairal2013stochastic}.

In this work, we focus on MM algorithms provably convergent to a critical point of Problem \ref{eq:generalprob} such as those described by \citet{kang2015global}.
Their main mechanism is to iteratively build a majorizing surrogate objective function that is easier to solve than the original learning problem.
In the case of non-convex penalties that are either fully concave or that can be written as the sum of a convex and concave functions, the idea is to linearize the concave part, and the next iterate is obtained by optimizing the resulting surrogate function.
Since $r_{\lambda}(|\cdot|)$ is a concave and differentiable function on $[0,+\infty)$, at any iterate $k$, we have :
$$
r_{\lambda}(|w_j|) \leq r_{\lambda}(|w_j^k|) + r_{\lambda}'(|w_j^k|) (|w_j| - |w_j^k|) \enspace.
$$
To take advantage of MM convergence properties \citep{kang2015global},
we also majorize the objective function and our algorithm boils down to the following iterative process
\begin{eqnarray}
\label{eq:mmalgo}
\w^{k+1} = \argmin_{\w \in \R^d} &  \frac{1}{2} \|\y - \X\w\|_2^2 +\frac{1}{2\alpha}\|\w - \w^k\|_2^2 \\ \nonumber& + \displaystyle\sum_{j=1}^d r_{\lambda}'(|w_j^k|)|w_j| \enspace, \nonumber
\end{eqnarray}
where $\alpha >0$ is some user-defined parameter controlling the proximal regularization strength.
Note that as $\alpha \rightarrow \infty$, the above problem recovers the reweighted $\ell_1$ iterations investigated by \citet{candes2008enhancing,gasso2009recovering}.
Moreover, when using $\w^k = \mathbb{0}$ (\eg when evaluating the first $\lambda$ in a path-wise fashion, for $k=0$) this recovers the Elastic-net penalty \cite{Zou_Hastie05}.


\section{\pwl: coordinate descent and screening}
\label{sec:wlasso}

As we have stated, the screening rule we have developed for regression with non-convex sparsity enforcing penalties is based on iterative minimization
of \pwl problem.
Hereafter, we briefly show that such sub-problems can be solved by iteratively optimizing coordinate wise.
Then, we derive a duality gap based screening rule to screen coordinate-wise.
In what follows, we assume that $\frac{0}{0}=0$\footnote{This will be of interest cases where $\lambda_j=0$ in Problem \ref{eq:weighted}.}.

\subsection{Primal and dual problems}
\label{sub:primal_dual_pb}

Solving the following primal problem (as it encompass
Problem \ref{eq:mmalgo} as a special case) would prove useful in the application of our MM framework:
\begin{equation}
  \label{eq:weighted}
  \min_{\w\in\R^d} P_{\Lambda}(\w) \triangleq \tfrac{1}{2} \| \y - \X \w\|_2^2 + \tfrac{1}{2\alpha} \|\w - \w^\prime\|_2^2 + \sum_{j=1}^d \lambda_j |w_j| \enspace.
\end{equation}
with $\w^\prime$ is some pre-defined vector of $\R^d$, and $\Lambda=(\lambda_1 \dots\lambda_d)^\top$ with $\lambda_1\geq0,\dots,\lambda_d \geq 0$ some regularization parameters.
In the sequel we denote Problem \ref{eq:weighted} as the \pwl (PWL) problem.

Problem \eqref{eq:weighted} can be solved through proximal gradient descent algorithm \cite{beck2009fast} but in order to benefit from screening rules, coordinate descent algorithms are to be privileged.
\citet{friedman2010regularization} have proposed a coordinate-wise update for the Elastic-net penalty and similarly, it can be shown that for the above, the following update holds for any coordinate $w_j$ and $\lambda_j \geq 0$:
\begin{align} \label{eq:cdupdate}
    w_{j} \leftarrow \frac{1}{\|\x_j\|_2^2 + \frac{1}{\alpha}} \sign(t_j) \max(0,|t_j| - \lambda_j) \enspace,
\end{align}
with $t_j =\x_j^\top (\y- \X\w + \x_j w_j) + \frac{1}{\alpha} w_{j}^\prime$.

Typically, coordinate descent algorithms visit all the variables in
a cyclic way (another popular choice is sampling uniformly at random among the coordinates) unless some screening rules prevent them from unnecessary updates.

Recent efficient screening methods rely on producing primal-dual approximate solutions and on defining tests based on these approximate solutions \cite{fercoq2015gap,shibagaki2016simultaneous,Ndiaye_Fercoq_Gramfort_Salmon16b}.
While our approach follows the same road, it does not result from a straightforward extension of the works of \citet{fercoq2015gap} and \citet{shibagaki2016simultaneous} due to the proximal regularizer.

The primal objective of \pwl (given in \Cref{eq:weighted}) is convex (and lower bounded) and admits at least one global solution.
To derive our screening tests, we need to investigate the dual formulation associated to this problem, that reads:
\begin{align} \label{eq:dual}
  \max_{\substack{\s \in \R^n\\ \v \in \R^d}}  D(\s,\v) &\triangleq -\frac{1}{2} \|\s\|_2^2 -\frac{\alpha}{2}\|\v\|_2^2  + \s^\top \y - \v^\top \w^\prime \\[-0.21cm]
&\text{s.t.} \quad |\X^\top \s - \v | \preccurlyeq \Lambda \enspace, \label{eq:dualconstraint}
 \end{align}
 with the inequality operator $\preccurlyeq$ being applied in a coordinate-wise manner.
As a side result of the dual derivation (see Appendix for the details), key conditions for screening any primal variable $w_j$ are obtained as:
$$
|\x_j^\top\s^\star - v_j^\star| - \lambda_j < 0 \implies  w_j^\star = 0\enspace,
$$
with $\s^\star$ and $\v^\star$ being solutions of the dual formulation from \Cref{eq:dual}.

\subsection{Screening test}
\label{sub:screening_test}

Screening test on the \pwl problem can thus be derived if we are able
to provide an upper bound on $|\x_j^\top\s^\star - v_j^\star|$ that is guaranteed to be strictly smaller than $\lambda_j$.
Suppose that we have an intermediate triplet of primal-dual solution
$(\hat \w, \hat \s, \hat \v)$ with $\hat \s$ and $\hat \v$ being dual feasible\footnote{meaning $|\X^\top \hat\s - \hat\v | \preccurlyeq \Lambda$.}, then
we can derive the following bound
\begin{align}\label{eq:bound}
  |\x_j^\top \s^\star - v^\star_j| &=|\x_j^\top \hat \s - \hat v_j + \x_j^\top (\s^\star - \hat \s) - (v_j^\star - \hat v_j)| \nonumber\\
&\leq|\x_j^\top \hat \s - \hat v_j|  + \|\x_j\| \|\s^\star - \hat \s\| + |v_j^\star - \hat v_j| \enspace.\nonumber
\end{align}

Now we need an upper bound on the distance between the approximated and the optimal dual solution in order to make screening condition exploitable. By exploiting the property that the objective function
$D(\s,\v)$ of the dual problem given in Equation \eqref{eq:dual} is quadratic and strongly concave\footnote{see \citep{Nesterov04} for a precise definition of strong convexity/concavity.}, the following inequality holds
\begin{align*}
D(\hat \s, \hat \v) \leq & D(\s^\star,\v^\star) - \nabla_\s D(\s^\star,\v^\star)^\top
(\hat \s - \s^\star)  \\
& - \nabla_\v D(\s^\star,\v^\star)^\top
(\hat \v - \v^\star)\\ & - \frac{1}{2} \| \hat \s - \s^\star\|_2^2
- \frac{\alpha}{2} \| \hat \v - \v^\star\|_2^2 \enspace,
\end{align*}
with $\nabla_\s D = \Big[\frac{\partial D}{\partial s_1}, \dots, \frac{\partial D}{\partial s_n}\Big]^\top$,  $\nabla_\v D = \Big[\frac{\partial D}{\partial v_1}, \dots, \frac{\partial D}{\partial v_d}\Big]^\top$.
As the dual problem is a constrained optimization problem, the first-order optimality condition for $(\s^\star,\v^\star)$ reads $\nabla_\s D(\s^\star,\v^\star)^\top( \s - \s^\star)  + \nabla_\v D(\s^\star,\v^\star)^\top ( \v - \v^\star) \geq 0 $, $\forall \s \in \R^n, \v\in \R^d$; thus we have
$$
2 (D(\s^\star,\v^\star) - D(\hat \s,\hat \v) ) \geq \| \hat \s - \s^\star\|_2^2
+ \alpha \|\hat \v - \v^\star\|_2^2 \enspace.
$$
By strong duality, we have $P_{\Lambda}(\hat \w)\geq D(\s^\star,\v^\star)$, hence
$${2 (P_{\Lambda}(\hat \w) - D(\hat \s, \hat \v))} \geq \| \hat \s - \s^\star\|_2^2 + \alpha \|\hat \v - \v^\star\|_2^2 \enspace.
$$
We can now use the duality gap for bounding $\| \hat \s - \s^\star\|$ and $|\hat v_j - v_j^\star|$.
Hence, given a primal-dual intermediate solution  $(\hat \w, \hat \s, \hat \v)$, with duality gap $G_{\Lambda}(\hat \w, \hat \s, \hat \v)\triangleq P_{\Lambda}(\hat \w) - D(\hat \s, \hat \v)$, the screening test for a variable $j$ is
\begin{equation}\label{eq:test}
  \underbrace{|\x_j^\top \hat \s - \hat v_j|  +  \sqrt{2 G_{\Lambda}(\hat \w, \hat \s, \hat \v)} \Big (\|\x_j\| + \frac{1}{\alpha} \Big)}_{T^{(\lambda_j)}_j(\hat \w, \hat \s, \hat \v)} < \lambda_j \enspace,
\end{equation}
and we can safely state that the $j$-th coordinate of $\w^\star$ is zero when this happens.

\paragraph{Finding approximate primal-dual solutions:}
In our case of interest, the \pwl problem is solved in its primal form using a coordinate descent algorithm that optimizes one coordinate at a time.
Hence, an approximate primal solution $\hat \w$ is easy to obtain by considering the current solution at a given iteration of the algorithm.
From this primal solution, we show how to obtain a dual feasible solution $\hat \s$ that can be considered for the screening test.

One can check the following primal/dual link (for instance by deriving the first order conditions of the maximization in Equation \ref{eq:where_the_link_come_from} or equivalent formulation \ref{eq:rewrite_the_link_problem}, see Appendix)
$$
\y - \X \w^\star= \s^\star \quad \text{ and } \quad \w^\star - \w^{\prime\star} = \alpha \v^\star
$$
with the constraints that
$
|\x_j^\top \s^\star - v_j^\star| \leq \lambda_j, \forall j \in [d].
$

Hence, a good approximation of the dual solution can be obtained
by scaling the residual vector $\y - \X \hat\w$ such that it becomes dual feasible.
Indeed, the condition $|\x_j^\top \s^\star - v_j^\star| \leq \lambda_j, \forall j \in[d]$ is guaranteed only at optimality.
To avoid issues with dividing by vanishing $\lambda_j$'s, let us consider the set $\mathcal{S} = \{j\in [d]: \lambda_j >0\}$ of associated indexes and assume that this set is non-empty.
Then, one can define
\begin{equation}\label{eq:jdagger}
j^\dagger = \argmax_{j \in \mathcal{S}}
\underbrace{\tfrac{1}{\lambda_j}\left|\x_j^\top (\y - \X \hat \w) - \tfrac{1}{\alpha}(\hat w_j - w_j^\prime)\right|}_{\rho(j)}\enspace.
\end{equation}
For all $j \in \mathcal{S}$, if $\rho(j^\dagger) \leq 1$ then , $\hat \s \triangleq \y - \X \hat \w$ and $\v = \tfrac{1}{\alpha}(\hat w - w^\prime)$ are dual feasible, no scaling is needed.
If $\rho(j^\dagger) > 1$, we define the approximate dual solution as $\hat \s = \frac{\y - \X\hat\w}{\rho(j^\dagger)}$ and $\hat \v = \frac{\hat \w - \w^\prime}{\alpha \rho(j^\dagger)}$ which are dual feasible.
Hence, in practice, we compute our screening test using the triplet
 $\{\hat \w, \frac{\y - \X\hat\w}
{\max(1,\rho(j^\dagger))}, \frac{\hat \w - \w^\prime}{\alpha \max(1,\rho(j^\dagger))}\}$.
\begin{remark}
As the above dual approximation is valid only for components of $\lambda_j >0$, we have added a special treatment for components $j \in [d] \setminus \mathcal{S}$ setting $\hat v_j = \x_j^\top \hat \s$ for such indexes, guaranteeing dual feasibility.
Note that the special case $\lambda_j = 0$ is not an innocuous case.
For some non-convex penalties like MCP or SCAD, their gradient $r_{\lambda}^\prime$ is equal to $0$ for large values and this is one of their key statistical property.
Hence, the situation in which $\lambda_j = 0$ is very likely to occur in practice for these penalties.
\end{remark}

\begin{algorithm}[t]
	\caption{Proximal Weighted Lasso (PWL) \label{alg:PWL} }
	\label{alg:pwl}
	\begin{algorithmic}[1]
		\REQUIRE{$\X$, $\y$, $\w^0$,$\Lambda$, $\alpha$, LstScreen}
		\ENSURE{$\w$, $\r = \y - \X\w$}
		\STATE $k \gets 0$
				\REPEAT
		\FOR {variable $j \not \in $ LstScreen}
		\STATE $w_j^{k+1} \leftarrow$ update coordinate $w_j^k$ using Equation \eqref{eq:cdupdate}
		\ENDFOR
		\STATE compute duality gap, $\r=\y - \X\w^{k+1}$, approximate dual variables
		\FOR {variable $j \not \in $ LstScreen}
		\STATE update screening condition using Equation \eqref{eq:test}
		\ENDFOR
		\STATE $k\leftarrow k +1$
		\UNTIL {convergence}
	\end{algorithmic}
\end{algorithm}

\paragraph{Comparing with other screening tests:}

Our screening test exploits duality gap and strong concavity of
the dual function, and the resulting test is similar to the one derived
by \citet{fercoq2015gap}.
Indeed, it can be shown that Equation \ref{eq:test} boils down to be a test based on a sphere centered on $\hat s$ with radius $\sqrt{2G_{\Lambda}(\hat \w, \hat \s,\hat \v)}$.
The main difference relies on the threshold of the test $\lambda_j$, which differs on a coordinate basis instead of being uniform.

Similarly to  the GAP test of \citet{fercoq2015gap}, our screening rule come with theoretical properties.
It can be shown that if we have a converging sequence $\{\hat \w^k\}$ of primal coefficients  $\lim_{k \rightarrow \infty} \hat \w^k = \w^\star$, then $\hat \s^k$ and $\hat \v^k$ defined as above converge to $\s^\star$ and $\v^\star$.
Moreover, we can state a property showing the ability of our screening rule to remove irrelevant variables after a finite number of iterations.

\begin{property}
Define the \textbf{Equicorrelation set}\footnote{following a terminology introduced by\citet{Tibshirani13}} of the \pwl as $ {\cE^\star} \triangleq \{j \in [d] : |\x_j^\top \s^\star - {v_j^\star}| = \lambda_j \} $ and $ \cE_k \triangleq \{j \in [d] : |\x_j^\top {\hat \s_k} -  {\hat v_j^k}| \geq \lambda_j \}$ obtained at iteration $k$
of an algorithm solving the \pwl.
Then, there exists an iteration $k_0\in\mathcal{N}$ s.t. $\forall k \geq k_0$, $\cE_k \subset {\cE^\star}$.
\end{property}

\begin{proof}
Because $\hat \w^k$, $\hat \s^k$ and $\hat \v^k$ are convergent, owing to the strong duality, the duality gap also converges towards zero.
Now, for any given $\epsilon$, define $k_0$ such that $\forall k \geq k_0$, we have
$$
\|\hat \s^k - \s^\star\|_2 \leq \epsilon, \|\hat \v^k - \v^\star\|_\infty \leq \epsilon \text{~and~} \sqrt{2G_\Lambda} \leq \epsilon\enspace.
$$

For $j \not \in \cE^\star$, we have
\begin{align}\nonumber
|\x_j^\top(\hat \s^k - \s^\star) - (\hat v^k_j - v_j^\star)| &\leq
|\x_j^\top(\hat \s^k - \s^\star)|+ |(\hat v^k_j - v_j^\star)| \\\nonumber
&\leq (\max_{j \not \in \cE^\star} \|{\x_j}\| + 1)\epsilon \enspace.
\end{align}
From triangle inequality we have:
\begin{align}\nonumber
|\x_j^\top\hat\s^k  - \hat v^k_j| &\leq |\x_j^\top(\hat\s^k - \s^\star) - (\hat v^k_j - v_j^\star)| + |\x_j^\top\s^\star  - v_j^\star| \\\nonumber
&\leq  (\max_{j \not \in \cE^\star} \|{\x_j}\| + 1)\epsilon +  |\x_j^\top\s^\star  - v_j^\star|
\end{align}
If we add $\sqrt{2G_\Lambda} {\Big (\|\x_j\| + \frac{1}{\alpha} \Big)}$ on both sides, we get
$$
T_j^{(\lambda_j)} \leq  \left(2\max_{j \not \in \cE^\star} \|{\x_j}\| + 1 + \frac{1}{\alpha} \right)\epsilon +   |\x_j^\top\s^\star  - v_j^\star|\enspace.
$$
Now define the constant $C \triangleq \min_{j \not \in \cE^\star} [\lambda_j - |\x_j^\top \s^\star - v_j^\star|]$ and because $j \not \in \cE^\star$, $C>0$.
Hence, if we choose
$$\epsilon < \frac{C}{  2\max_{j \not \in \cE^\star} \|{\x_j}\| + 1 + \frac{1}{\alpha}} \enspace,
$$
we have $T_j^{(\lambda_j)} < {\lambda_j} $ which means that the variable $j$ has been screened, hence $j \not \in \cE_k$.
To conclude, we have $j \not \in \cE^\star$ implies that $j \not \in \cE_k$, which also translates in $\cE_k \subset \cE^\star$.
\end{proof}
This property thus tells that  all the zero variables of $\w^\star$ are correctly detected and screened  by our screening rule in a finite number of iterations of the algorithm.

\section{Screening rule for non-convex regularizers}
\label{sec:screening_rule_for_non_convex_regularizer}

Now that we have described the inner solver and its screening rule, we are going to analyze how this rule can be improved into a majorization-minimization (MM) approach.

\subsection{Majorization-minimization approaches and screening}
\label{sub:majorization_minimization_approaches_and_screening}

At first, let us discuss some properties of our MM algorithm.
According to the first order condition of the \pwl related to the MM problem at iteration $k$, the following inequality holds for any $j \in[d]$
$$
|\x_j^\top \s^{k,\star} - v_{j}^{k,\star}| \leq \lambda_{j}^k\enspace,
$$
where the superscript denotes the optimal solution at iteration $k$.
Owing to the convergence properties of the MM algorithm \citet{kang2015global}, we know that the sequence $\{\w^k\}$ converges towards a vector satisfying
\Cref{eq:fermat}.
Owing to the continuity of $r_{\lambda}(|w|)$, we deduce that the sequence $\{\lambda_j^k\}$ also converges towards a $\lambda_{j}^\star$.
Thus, by taking the limits of the above inequality, the following condition holds for $w_j^\star=0$:
$$
|\x_j^\top \s^{\star} - v_{j}^{\star}| \leq \lambda_{j}^\star\enspace,
$$
with $\lambda_j^\star =  r_{\lambda}'(|w_j^\star|)$.
This inequality basically tells us about the relation between vanishing primal component and the optimal dual variable at each iteration
$k$.
While this suggests that screening within each inner problem by defining $\lambda_j$ in \Cref{eq:weighted} as $\lambda_j^k =  r_{\lambda}'(|w_j^k|)$  should improve efficiency of the global MM solver, it does not tell whether screened variables at iteration $k$ are going to be screened at the next iteration as the $\lambda_j^k$'s are also expected to vary between iterations.

\begin{remark}
In general, the behavior of a $\lambda_j$ across MM iterations strongly depends on an initial $\w^0$ and the optimal $\w^\star$.
For instance, if one variable $w_j^0$ is initialized at $0$ and $w_j^\star$ is large, $\lambda_j$ will tend to be decreasing across iterations.
Conversely, $\lambda_{j}$ will tend to increase.
\end{remark}

\subsection{Propagating screening conditions}
\label{sub:propagating_screening_conditions}

In what follows, we derive conditions on allowing to propagate screened coefficients from one iteration to another in the MM framework.

\begin{property}
Consider a \pwl problem with weights $\{\lambda_j\}$ and its primal-dual approximate solutions $\hat{\w}$, $\hat \s$ and $\hat \v$ allowing to evaluate a screening test in Equation \ref{eq:test}.
Suppose that we have a new set of weight $\Lambda^{\newexp}=\{\lambda_j^{\newexp}\}_{j=1,\dots,d}$ defining a new \pwl problem.
Given a primal-dual approximate solution defined by the triplet ($\hat \w^{\newexp}$,$\hat \s^{\newexp}, \hat \v^{\newexp}$) for the latter problem, a  screening test for variable $j$ reads
\begin{equation}\label{eq:new}
	T^{(\lambda_j)}_j(\hat \w, \hat \s, \hat \v) + \|\x_j\|(a + \sqrt{2b}) + c + \frac{1}{\alpha}\sqrt{2b} < \lambda_j^{\newexp} \enspace,
\end{equation}
where $T^{(\lambda_j)}_j(\hat \w, \hat \s, \hat \v)$ is the screening test for $j$ at $\lambda_j$, $a$, $b$ and $c$ are constants such that $\|\hat \s^{\newexp} - \hat \s\|_2 \leq a$,
$|G_{\Lambda}(\hat \w, \hat \s, \hat \v) - G_{\Lambda^{\newexp}}(\hat \w^{\newexp}, \hat \s^{\newexp}, \hat \v^{\newexp})| \leq b$
and $|\hat v_j^{\newexp} - \hat v_j| \leq c$.
\end{property}

\begin{proof}
The screening test for the novel \pwl problem for parameter $\Lambda_j^{\newexp}$ can be written as
\begin{equation*} \label{eq:test2}
	|\x_j^\top \hat \s^{\newexp} - \hat v^{\newexp}|  +  \sqrt{2 G_{\Lambda^{\newexp}}(\hat \w^{\newexp}, \hat \s^{\newexp}, \v^{\newexp})} \left(\|\x_j\|  + \tfrac{1}{\alpha}\right)< \lambda_j^{\newexp} \enspace.
\end{equation*}
Let us bound the terms in the left-hand side of this inequality. At first, we have:
\begin{align}\nonumber
|\x_j^\top \hat \s^{\newexp} - \hat v^{\newexp}|  &\leq |\x_j^\top \hat \s - \hat v_j| + |\x_j^\top (\hat \s^{\newexp} - \hat \s)| + |\hat v_j^{\newexp} - \hat v_j|\\\nonumber
& \leq  |\x_j^\top \hat \s - \hat v_j| + \|\x_j\| \|\hat \s^{\newexp} - \hat \s\| + |\hat v_j^{\newexp} - \hat v_j| \enspace, \nonumber
\end{align}
and
\begin{align}
	\sqrt{G_{\Lambda^{\newexp}}} & \leq \sqrt{|G_{\Lambda^{\newexp}} - G_{\Lambda}| + |G_{\Lambda}|}\\\nonumber
	&\leq \sqrt{|G_{\Lambda^{\newexp}} - G_{\Lambda}|} + \sqrt{|G_{\Lambda}|} \enspace.
\end{align}
where the last inequality holds by applying the norm property $\|\x\|_2 \leq \|\x\|_1$
to the 2-dimensional vector of component  $[\sqrt{|G_{\Lambda^{\newexp}} - G_{\Lambda}|},\sqrt{|G_{\Lambda}|}]$, where we have drop the dependence on $(\hat \w, \hat \s, \hat \v)$ and $(\hat \w^{\newexp}, \hat \s^{\newexp}, \hat \v^{\newexp})$ for simplicity.
By gathering the pieces together, the left-hand side of Equation \ref{eq:test2} can be bounded by
\begin{align}
|\x_j^\top \hat \s - \hat v_j| + \|\x_j\|  \|\hat \s^{\newexp} - \hat \s\|  +  \sqrt{2G_{\Lambda}} \left(\|\x_j\| + \tfrac{1}{\alpha}\right) \\
 + \sqrt{2|G_{\Lambda^{\newexp}} - G_{\Lambda}|}\left(\|\x_j\| + \tfrac{1}{\alpha}\right) + |\hat v_j^{\newexp} - \hat v_j| \enspace. \nonumber
\end{align}
which leads us to the novel screening test
\begin{equation}
T^{(\lambda_j)}_j(\hat \w, \hat \s, \hat \v)
+ \|\x_j\| (a + \sqrt{2b}) + c + \tfrac{1}{\alpha} \sqrt{2b} \leq \lambda_j^{\newexp} \enspace.
\end{equation}
with $T^{(\lambda_j)}_j(\hat \w, \hat \s, \hat \v)$ defined in \Cref{eq:test}.
\end{proof}
In order to make this screening test tractable, we need at first an approximation $\hat \s^{\newexp}$ of the dual solution, then an upper bound on the norm of $\|\hat{\s} - \hat{\s}^{\newexp}\|$ and a
bound on the difference in duality gap $|G_{\Lambda^{\newexp}} - G_{\Lambda}|$. In practice, we will define $\hat{\s}^{\newexp} = \frac{\y - \X\hat\w}{\rho^{\newexp}(j^\dagger)}$ and then compute
exactly  $\|\hat{\s} - \hat{\s}^{\newexp}\|$ and $|G_{\Lambda^{\newexp}} - G_{\Lambda}|$.
Interestingly, since we consider the primal solution $\hat \w$ as our approximate primal solution for the new problem, computing $\rho^{\newexp}(j)$ is only costs an element-wise division since $\y - \X \hat\w$ and $|\x_j^\top(\y - \X \hat\w)|$ have already been pre-computed at the previous MM iteration.
Another interesting point to highlight is that given pre-computed screening values $\{T^{(\lambda_j)}_j(\hat \w, \hat \s, \hat \v)\}$, the screening test given in \Cref{eq:new} does not involve any additional dot product and thus is cheaper to compute.

\begin{figure*}[t]
	~\hfill\includegraphics[width=8cm]{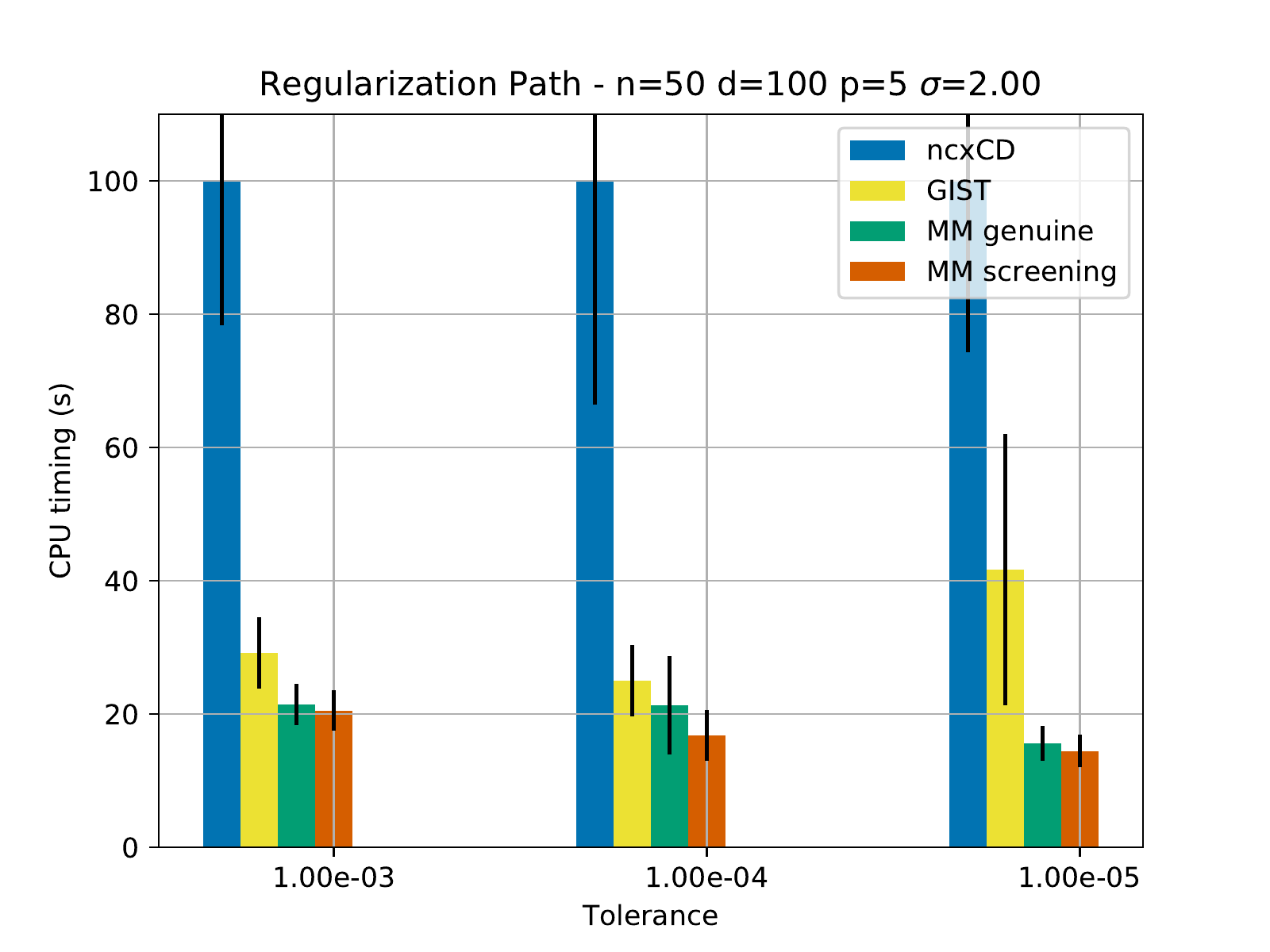}~\hfill~
	\includegraphics[width=8cm]{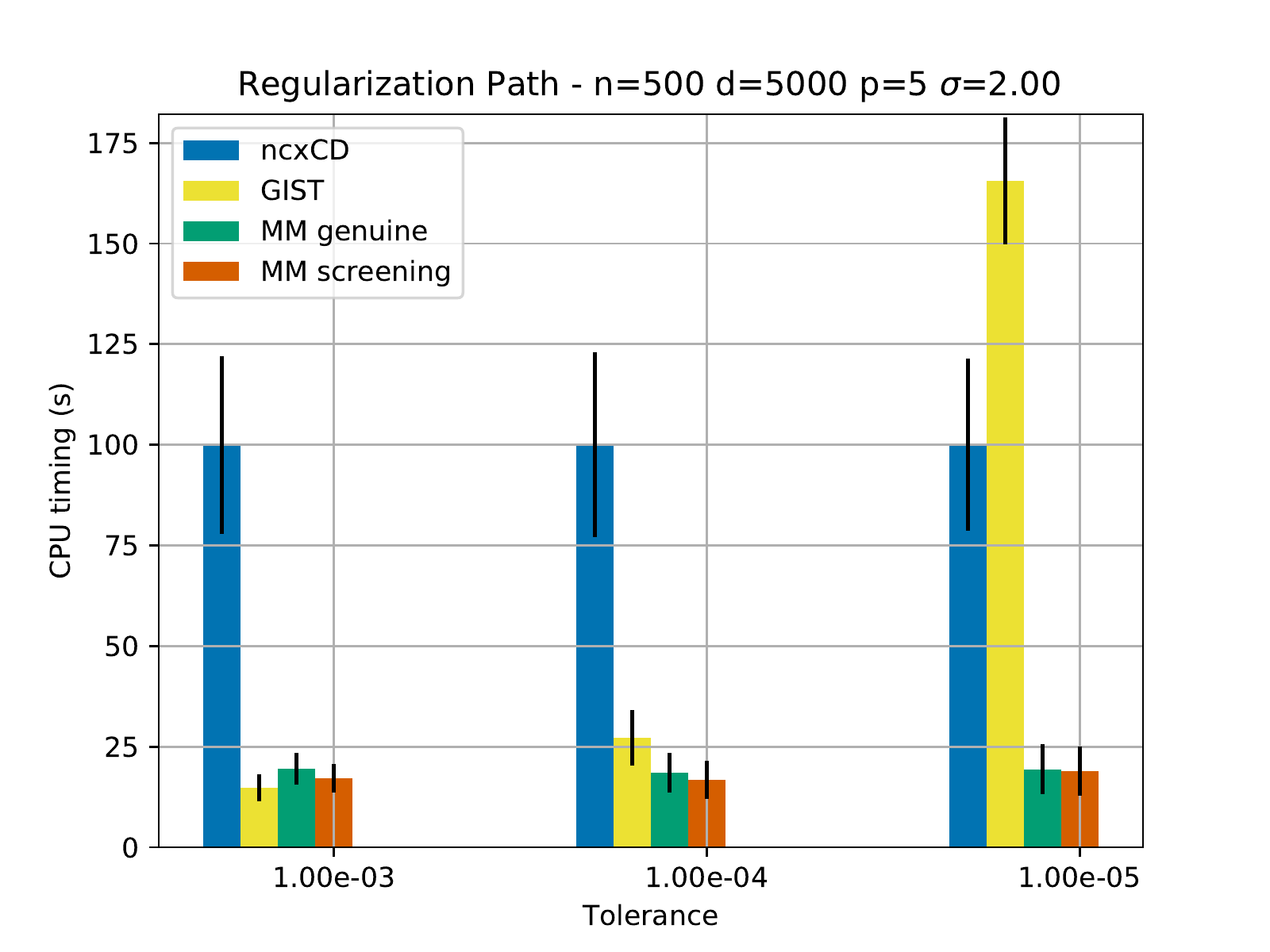}~\hfill
			\caption{Comparing running time for coordinate descent (CD) and proximal gradient algorithms as well as screening-based CD and MM algorithms under different tolerances on the stopping condition and under different samples, features settings. For all experiments, we have $5$ active variables. (left) $n=50$,
		$d=100$, $\sigma=2$. (right) $n=500$,
		$d=5000$, $\sigma=2$
		\label{fig:regpathtoy}}
\end{figure*}

\subsection{Algorithm}
\label{sub:algorithms}

\begin{algorithm}[t]
		\caption{MM algorithm for Lasso with penalty $r_{\lambda}(|w|)$}
		\label{alg:mm_screen}
	\begin{algorithmic}[1]
		\REQUIRE{$\X$, $\y$, $\lambda$, $\w^0$, $\alpha$}
		\STATE $k \gets 0$
				\REPEAT
		\STATE $\Lambda^k=\{\lambda_j^k\}_{j=1,\dots,d} \gets \{ r'_{\lambda}(|w_j^k|)\}_{j=1,\dots,d}$
		\STATE compute approximate dual $\s^k$ and duality gap $G_{\Lambda^k}$ given $\w^k$ and $\Lambda^k$
		\IF {needed}
		\STATE compute screening scores $T^{(\lambda_j)}_j(\w^k, \s^k, \v^k)$ according to Eq.~\eqref{eq:test}
		\STATE store reference duality gap and approximate dual
		\ELSE
		\STATE estimate screening scores according to Eq.~\eqref{eq:new}
		\ENDIF
		\STATE LstScreen $\gets$ updated screened variables list based on results of Line $6$ or $9$
		\STATE {\small $\w^{k+1}$, $\y - \X\w^{k+1}$ $\leftarrow $ PWL($\X$,$\y$,$\w^k$,$\Lambda^k$,$\alpha$,Lstcreen)}
		\STATE $k \leftarrow k + 1$
		\UNTIL {convergence}
	\end{algorithmic}
\end{algorithm}

When considering MM algorithms for handling non-convex penalties, we advocate the use of weighted Lasso with screening as a solver for Equation \ref{eq:mmalgo} and a screening variable propagation condition as given in Equation \ref{eq:new}.
This last property allows us  to cheaply evaluating whether a variable can be screened before entering in the inner solver.
However, \Cref{eq:new} also needs the screening score computed at some previous $\{\lambda_j\}$ and a trade-off has thus to be sought between relevance of the test when $\lambda_j^\nu$ is not too far from $\lambda_j$ and the computational time needed for evaluating
$T^{(\lambda_j)}_j$.
In practice, we compute the exact score $T^{(\lambda_j)}_j(\hat \w, \hat \s, \hat \v)$ every $10$ iterations and apply \Cref{eq:new} for the rest of the iterations.
This results in \Cref{alg:mm_screen} where the inner solvers, denoted as PWL and solved according to Algorithm \ref{alg:PWL}, are warm-started with previous outputs and provided with a list of already screened variables.
In practice, $\alpha>0$ helps us guaranteeing theoretical convergence of the sequence $\{\w^k\}$ and we have set $\alpha= 10^9$ for all our experiments, leading to very small regularization allowing large deviations from $\w^k$.


\section{Numerical experiments}
\label{sec:numerical_experiments}

The goal of screening rules is to improve the efficiency of solvers by focusing
only on variables that are non-vanishing. In this section, we thus report the
computational gains we obtain owing to our screening strategy.

\subsection{Experimental set-up}

\begin{figure*}[t]
	\centering
	\includegraphics[width=5cm]{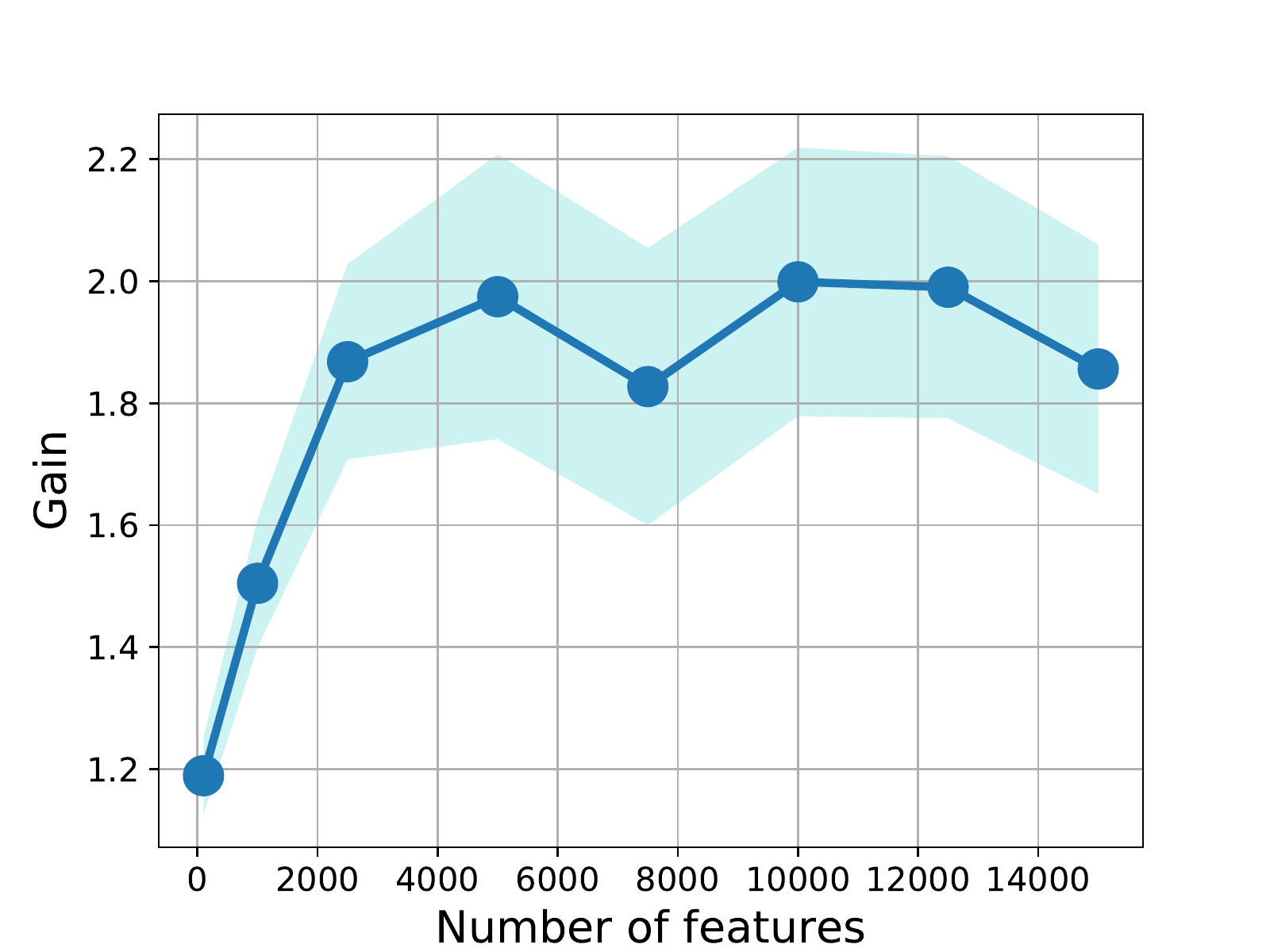}
	\includegraphics[width=5cm]{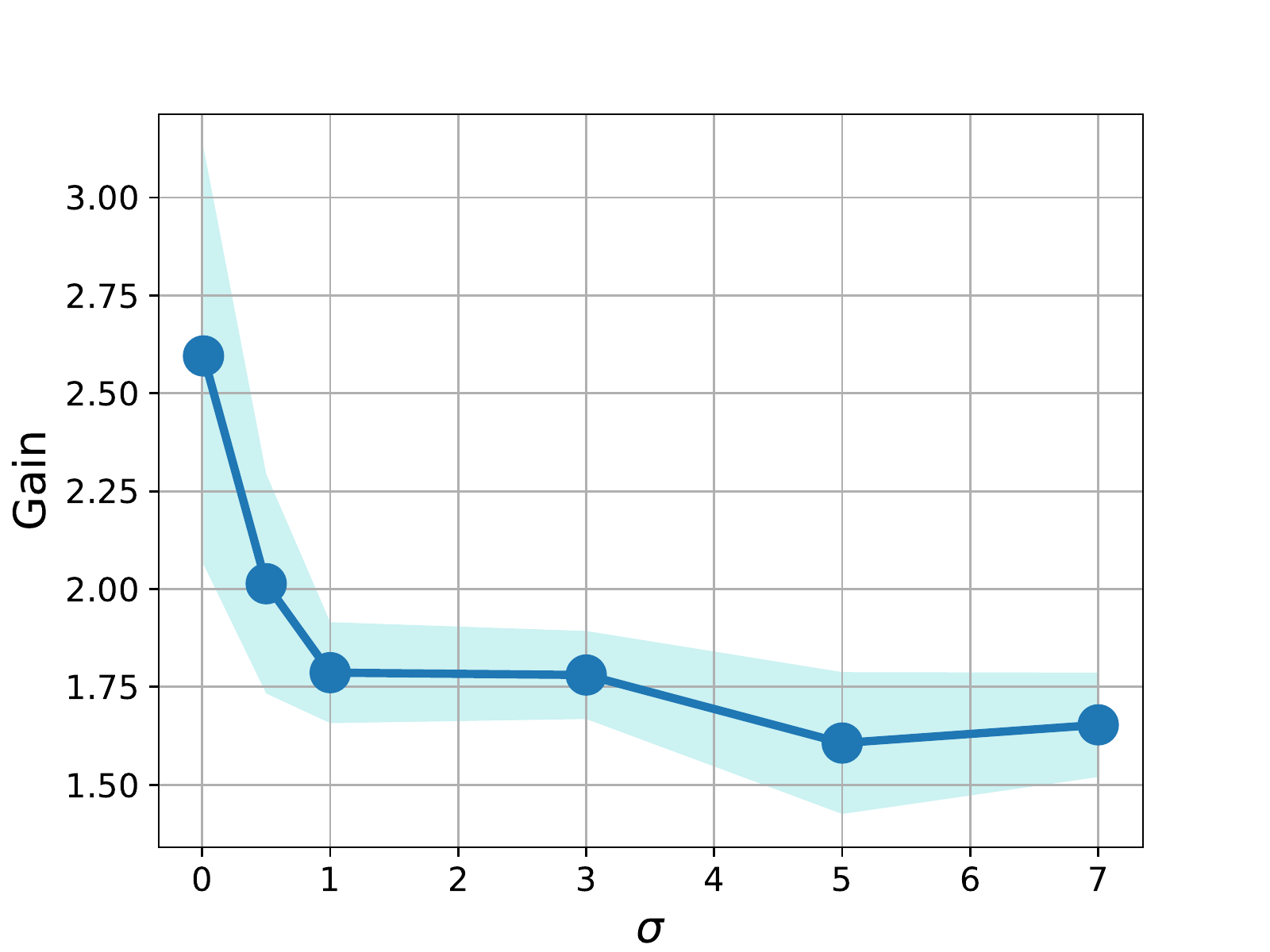}
	\includegraphics[width=5cm]{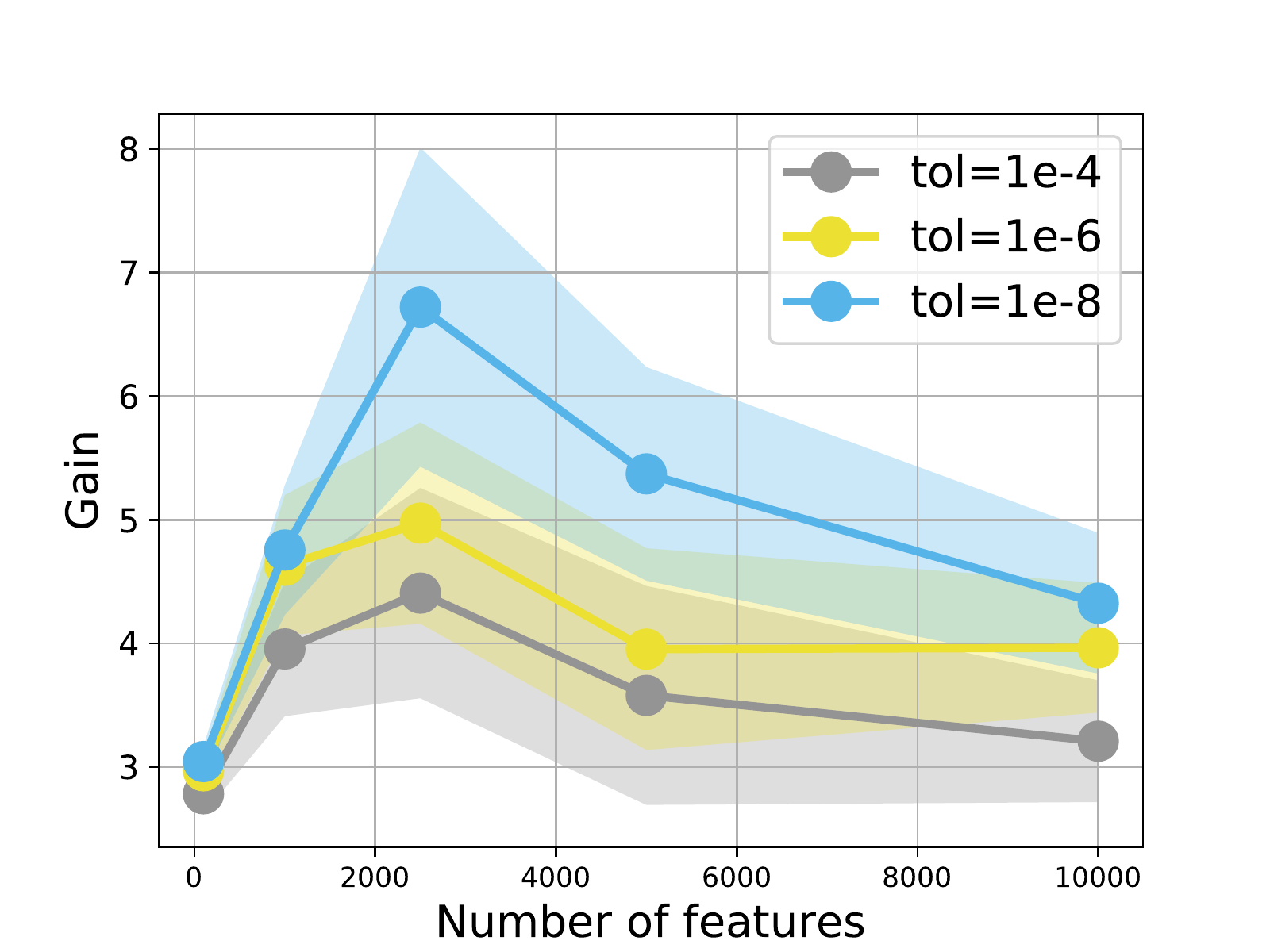}
	\caption{\label{fig:gain}Computational gain of propagating screening within MM iterations (left) \wrt number of features  at a KKT tolerance of  $10^{-4}$. (middle) \wrt to noise level at a KKT tolerance of  $10^{-4}$.
	(right) \wrt number of features at different tolerance in a low-noise regime.}
\end{figure*}
\begin{figure*}[t]
	~\hfill\includegraphics[width=6cm]{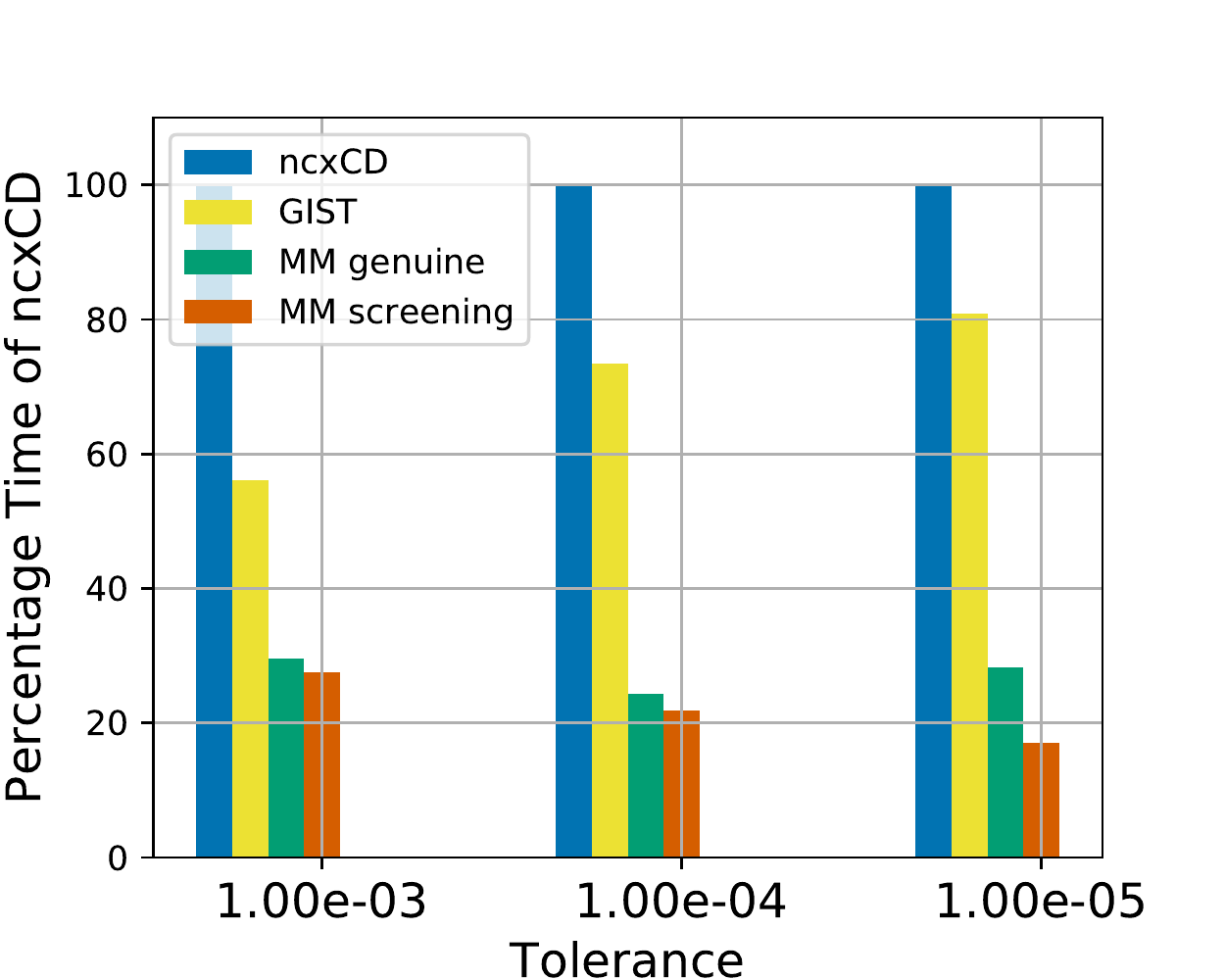}~\hfill~
	~\hfill\includegraphics[width=6cm]{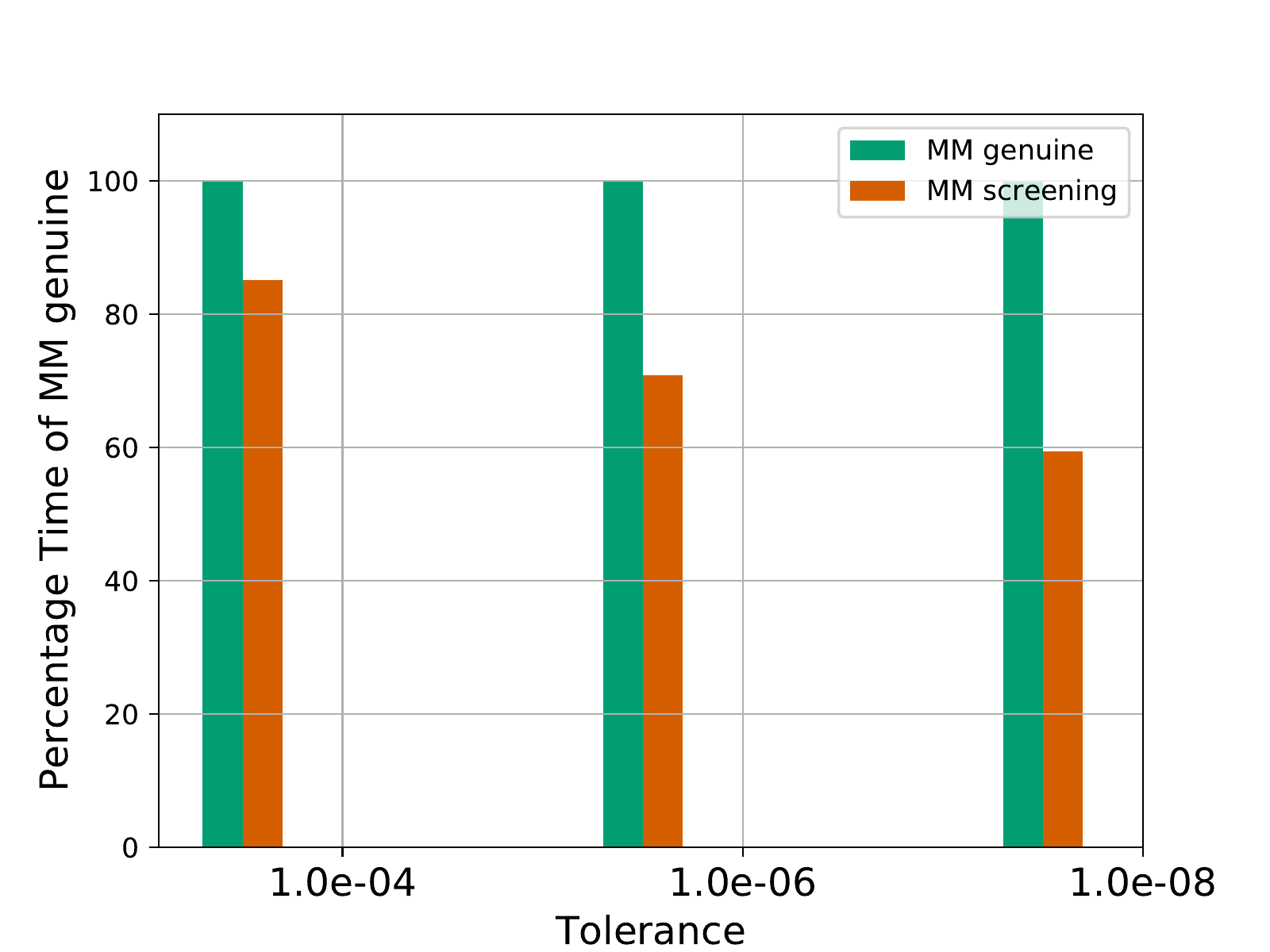}~\hfill~
	\caption{ Running time for computing the regularization path on (left) the dense Leukemia dataset $n=72$, $d=7129$. (right) sparse dataset Newsgroup with $n=961$ and $d=21319$. \label{fig:leuk}}
\end{figure*}

Our goal is to compare the computational running time of different algorithms
for computing a regularization path on problem \eqref{eq:generalprob}.
For most relevant non-convex regularizers, coordinatewise update \cite{breheny2011coordinate} and proximal operator \cite{gong2013general}
can be derived in a closed-form. For our experiments, we have used the log-sum penalty which has an hyperparameter
$\theta$. Hence, our regularization path involves $2$ parameters ($\lambda_t$,$\theta_t$). The set of $\{\lambda_t\}_{t=0}^{N_t-1}$ has been defined as $\lambda_t \triangleq
\lambda_{\max} 10^{-\frac{ 3t}{N_t-1}}$, $N_t$ being dependent of the problems and $\theta \in \{0.01, 0.1,1\}$.

Our baseline method should have been an MM algorithm, in which each subproblem as given in Equation \ref{eq:mmalgo} is solved using a coordinate descent algorithm without screening. However, due to its very poor running time, we have omitted its performance.
Two other competitors that directly
address the non-convex learning problems have instead been investigated: the first one, denoted as \textbf{GIST}, uses a majorization of the loss function and iterative shrinkage-thresholding \citep{gong2013general} while the second one, named \textbf{ncxCD} is a coordinate descent algorithm that directly handles non-convex penalties \cite{breheny2011coordinate,mazumder2011sparsenet}.
We have named \textbf{MM-screen} our method that screens within each \pwl and propagates screening scores while the genuine version drops the screening propagation (but screening inside inner solvers is kept), is denoted \textbf{MM-genuine}.
All algorithms have been stopped when optimality conditions as described in Equation \ref{eq:optcond} are satisfied up to the same tolerance $\tau$.
Note that all algorithms have been implemented in Python/Numpy.
Hence, this may give GIST a slight computational advantage over coordinate descent approaches that heavily benefit on loop efficiency of low level languages.
For our MM approaches, we stop the inner solver of the \pwl when the duality gap is smaller than $10^{-4}$ and screening is computed every $5$ iterations.
In the outer iterations, we perform the screening every $10$ iterations.

\subsection{Toy problem}
\label{sub:toy_problem}

Our toy regression problem has been built as follows. The entries of the regression design matrix $\X \in \R^{n \times d}$
are drawn uniformly from a Gaussian distribution of zero mean and
variance $4$. For a given $n$ and $d$ and a number {$p$} of active
variables, the true coefficient vector $\w^\star$ is
obtained as follows. The {$p$} non-zero positions are chosen
randomly, and their values are drawn from a zero-mean
unit variance Gaussian distribution, to which we added $\pm 0.1$
according to the sign of $w_j^\star$.
Finally, the target vector is obtained as $\y = \X\w^\star + \e$ where $e$ is a random noise vector drawn from a Gaussian distribution with zero-mean and standard deviation $\sigma$. For the toy problem, we have set $N_t=50$.

Figure \ref{fig:regpathtoy} presents the running time needed for the
different algorithms to reach convergence under different settings. We note that indifferently to the settings our screening rules help in reducing the computational time
by a factor of about $5$ compared to a plain non-convex coordinate
descent. The gain compared to \textbf{GIST} is mostly notable
for high-precision and noisy problems.

We have also analyzed the benefit brought by our screening propagation strategy. Figure \ref{fig:gain} presents the gain in computation time
when comparing \textbf{MM screening} and \textbf{MM genuine}.
As we have kept the number of relevant features to $5$, one can note that for a fixed amount of noise the gain is around $1.8$
for a wide range of dimensionality. Similarly, the gain is almost constant
for a large range of noise and the best benefit occurs in a low-noise setting. More interestingly, we have compared this gain for increasing
precision and for a low-noise situation $\sigma=0.01$, which is classical
noise level in the screening literature \cite{ndiaye2016gap,tibshirani2012strong}. We can note from the mostright panel of Figure \ref{fig:gain} that the more precision we require on the resolution of the learning problem, the more we benefit from screening propagation. For a tolerance of $10^{-8}$, the gain peaks at about $6.5$ whereas in most regimes of number of features, the gain is about $4$. Even for a lower tolerance, the genuine screening approach is
$4$ times slower than the full approach we propose.

\subsection{Real-world datasets}
\label{sec:real_world_datasets}
We have also run the comparison on different real datasets.  Figure
\ref{fig:leuk} presents the results obtained on the \emph{leukemia} dataset, which is a dense data with $n=72$ examples in dimension $d=7129$. For the path computation, the set of ${\lambda_t}$ has been fixed to $N_t=20$ elements.
Remark that the gain compared to \textbf{ncvxCD} varies from
$3$ to $5$ depending on the tolerance and is about $4$ compared
to $\textbf{GIST}$ at high tolerance.
On the right panel of Figure \ref{fig:leuk}, we compare the gain in running
time brought by screening propagation rule in a real world sparse dataset \emph{newsgroups}
in which we have kept only $2$ categories (\emph{religion} and \emph{graphics}) resulting in $n=961$ and $d=21319$. We can note that the gain is similar to what we have observed on the toy problem ranging in between $1.3$ and $1.8$.

\section{Conclusion}

We have presented the first screening rule strategy
that  handles sparsity-inducing non-convex regularizers. The approach we propose
is based on a majorization-minimization framework in which
each inner iteration solves a \pwl problem.
We introduced a screening rule for this learning problem and a
rule for propagating screened variables within MM iterations.
Interestingly, our screening rule for the weighted Lasso is able to
identify all the variables to be screened in a finite amount of time.
We have carried out several numerical experiments showing the benefits
of the proposed approach compared to methods directly handling the
non-convexity of the regularizers and illustrating the situation in which our propagating-screening rule helps in accelerating efficiency of
the solver.


\bibliographystyle{icml2019}
\clearpage
\onecolumn

\section*{Supplementary Material of ``Screening rules for Lasso with  non-convex sparse regularizers"}
\subsection*{Dual problem of the weighted Lasso minimization}
\label{sec:appendix_1}

Let recall the weighted lasso problem
\begin{equation*}
\min_{\w\in\R^d} \tfrac{1}{2} \| \y - \X \w\|_2^2 + \tfrac{1}{2\alpha} \|\w - \w^\prime\|_2^2 + \sum_{j=1}^d \lambda_j |w_j| \enspace.
\end{equation*}
This is Elastic-Net type problem and can be expressed as
\begin{equation*}
\min_{\w\in\R^d} \tfrac{1}{2} \| \tilde \y - \tilde \X \w\|_2^2  + \sum_{i=1}^d \lambda_j |w_j|,
\end{equation*}
where $ \tilde \y = \begin{bmatrix}
\y \\
\frac{\w^\prime}{ \sqrt{\alpha}}
\end{bmatrix} \in \mathbb{R}^{n+d}$ and $ \tilde \X  = \begin{bmatrix}
\X \\
\frac{\I}{ \sqrt{\alpha}}
\end{bmatrix} \in \mathbb{R}^{(n+d) \times d} \enspace.$

Let $\tilde \a_i = \tilde \X_{i,:}^\top$ and  $\phi_i(z_i) = \tfrac{1}{2} (\tilde y_i - z_i)^2$ being the quadratic loss function. Let its convex conjugate \citep{Boyd_Cvx_book} being $\phi_i^*(\eta_i) = \max_{z_i} \eta_i z_i - \phi_i(z_i)$ for a scalar $\eta_i$, which results in $\phi_i^*(\eta_i) = \tfrac{1}{2} \eta_i^2 + \eta_i \tilde{y}_i$. Note also that $\phi_i = \phi_i^{**}$ as $\phi_i$ is convex.

Following \citep{johnson2015blitz} we derive the dual of the weighted problem through these steps
\begin{align}
& \min_{\w\in\R^d} \tfrac{1}{2} \| \tilde \y - \tilde \X \w\|_2^2  + \sum_{i=1}^d \lambda_j |w_j| \nonumber \\
& = \min_{\w\in\R^d} \tfrac{1}{2} \sum_{i=1}^{n+d} (\tilde y_i - \tilde \a_i^\top \w)^2  + \sum_{i=1}^d \lambda_j |w_j| \nonumber \\
& = \min_{\w\in\R^d} \sum_{i=1}^{n+d} \phi_i(\tilde \a_i^\top \w)  + \sum_{i=1}^d \lambda_j |w_j| \nonumber \\
& = \min_{\w\in\R^d} \sum_{i=1}^{n+d} \phi_i^{**}(\tilde \a_i^\top \w)  + \sum_{i=1}^d \lambda_j |w_j| \nonumber \\
& = \min_{\w\in\R^d} \sum_{i=1}^{n+d} \max_{\eta_i} [(\tilde \a_i^\top \w) \eta_i  - \phi_i^{*}(\eta_i)]  + \sum_{i=1}^d \lambda_j |w_j| \label{eq:where_the_link_come_from} \\
& = \min_{\w\in\R^d} \max_{\bfeta \in \mathbb{R}^{n+d}} - \sum_{i=1}^{n+d} \phi_i^{*}(\eta_i) + \w^\top \tilde{\X}^\top \bfeta   + \sum_{i=1}^d \lambda_j |w_j| \nonumber \\
& =  \max_{\bfeta \in \mathbb{R}^{n+d}} - \sum_{i=1}^{n+d} \phi_i^{*}(\eta_i) + \min_{\w\in\R^d} \w^\top \tilde{\X}^\top \bfeta + \sum_{i=1}^d \lambda_j |w_j| \nonumber \\
& = \max_{\bfeta: |\tilde{\X}^\top \bfeta | \preccurlyeq \Lambda} - \tfrac{1}{2} \|\bfeta\|_2^2 - \bfeta^\top \tilde{\y} \enspace.
\label{eq:partial_dual_weighted_lasso}
\end{align}
The dual objective function is obtained by substituting the expression $\phi_i^*$ and using the optimality condition of the problem 

\begin{equation}
\label{eq:lin_lasso}
\min_{\w\in\R^d} \w^\top \tilde{\X}^\top \bfeta + \sum_{i=1}^d \lambda_j |w_j| \enspace.
\end{equation}

This problem is separable and optimality condition with respect to any $w_j$ is as follows, provided $g_j = \left(\tilde \X^\top \bfeta\right)_j$
\begin{equation}
\label{eq:opt_cdt_lin_lasso}
\begin{cases}
g_j + \lambda_j \text{sign}(w_j) = 0 & \text{if }  w_j \neq 0 \\
 \left |g_j \right| \leq \lambda_j & \text{if }  w_j = 0.
\end{cases}
\end{equation}
The latter condition implies the coordinate-wise inequality constraint $|\tilde{\X}^\top \bfeta| \preccurlyeq \Lambda$ with $\Lambda^\top = \begin{pmatrix}
\lambda_1, \dots,  \lambda_d
\end{pmatrix}$.  Also we can easily establish that $ g_j w_j + \lambda_j |w_j| = 0$. Hence the objective function in \Cref{eq:lin_lasso} vanishes.

Finally let us decompose the dual vector as $ \bfeta = \begin{bmatrix}
-\s \\
\sqrt{\alpha} \v
\end{bmatrix}$ where $\s \in \mathbb{R}^{n}$ and $\v \in \mathbb{R}^{d}$. Recalling the form of $\tilde \y$ and $\tilde{\X}$, it is easy to see that the dual problem (\ref{eq:partial_dual_weighted_lasso}) becomes
\begin{equation*}
\max_{\s, \v: |{\X}^\top \s - \v | \preccurlyeq \Lambda} - \tfrac{1}{2} \|\s\|_2^2 - \tfrac{\alpha}{2} \|\v\|_2^2 + \s^\top \y - \v^\top \w^\prime \enspace.
\end{equation*}
Also, from (\ref{eq:opt_cdt_lin_lasso}), it holds the screening conditions
\begin{equation}
|\x_j^\top \s -  v_j| < \lambda_j \implies w_j = 0 \enspace, \quad \forall j \in[d] \enspace,
\end{equation}
remind that $\x_j = \X_{:,j}$ is the $j^{\text{th}}$ covariate.
In addition,  the maximisation in \Cref{eq:where_the_link_come_from} takes the form
\begin{equation}
\label{eq:rewrite_the_link_problem}
\max_{\s, \v} - \tfrac{1}{2} \|\s\|_2^2 - \tfrac{\alpha}{2} \|\v\|_2^2 + \s^\top (\y - \X \w) + \v^\top (\w - \w^\prime)   \enspace.
\end{equation}

Thus given an optimal solution $\w^\star$, we may have $\s^\star = \y - \X\w^\star$ and $\alpha \v^\star = \w^\star - \w^\prime$, by deriving the first order optimality conditions of this maximization problem.

\end{document}